\documentclass[runningheads]{llncs}
\usepackage[misc] {ifsym}
\usepackage{cite}
\usepackage{amsmath,amssymb,amsfonts}
\usepackage{algorithmic}
\usepackage{graphicx}
\usepackage{textcomp}
\usepackage{xcolor}
\usepackage{graphicx}
\usepackage{booktabs}
\usepackage{multirow}
\usepackage{algorithm}
\usepackage{bm}
\usepackage{graphicx}
\begin{document}
\title{Tri-Transformer Hawkes Process: Three Heads are better than one\thanks{This work was supported by  the Science Foundation of China University of Petroleum Beijing (No.2462020YXZZ023).}}
%
%
\author{Zhi-yan Song\inst{1} \and
Jian-wei Liu\inst{1}\orcidID{0000-0002-0634-4408}\textsuperscript{(\Letter)}
\and Lu-ning Zhang\inst{1}\orcidID{0000-0002-7333-5050}
\and Ya-nan Han\inst{1}\orcidID{ 0000-0002-0695-5604}}
\authorrunning{Z. Song et al.}
%
\institute{Department of Automation, College of Information Science and Engineering, China University of Petroleum Beijing (CUP), Beijing, China\\
\email{liujw@cup.edu.cn}\\
}
\tocauthor
\toctitle
\maketitle              
\begin{abstract}
Most of the real world data we encounter are asynchronous event sequence, so the last decades have been characterized by the implementation of various point process into the field of social networks, electronic medical records and financial transactions. At the beginning, Hawkes process and its variants which can simulate simultaneously the self-triggering and mutual triggering patterns between different events in complex sequences in a clear and quantitative way are more popular. Later on, with the advances of neural network, neural Hawkes process has been proposed one after another, and gradually become a research hotspot. The proposal of the transformer Hawkes process (THP) has gained a huge performance improvement, so a new upsurge of the neural Hawkes process based on transformer is set off. However, THP does not make full use of the information of occurrence time and type of event in the asynchronous event sequence. It simply adds the encoding of event type conversion and the location encoding of time conversion to the source encoding. At the same time, the learner built from a single transformer will result in an inescapable learning bias. In order to mitigate these problems, we propose a tri-transformer Hawkes process (Tri-THP) model, in which the event and time information are added to the dot-product attention as auxiliary information to form a new multi-head attention. The effectiveness of the Tri-THP is proved by a series of well-designed experiments on both real world and synthetic data.

\end{abstract}
\section{Introduction}
Modeling and predicting the asynchronous event sequences are extensively utilized in different domains, such as financial data [1], genome analysis [2], earthquake sequences [3], electronic medical records [4], social network [5], etc. In order to obtain effective information from these asynchronous event data, analyze the relationship between events and predict the events that may occur in the future, the most common and effective measure is the point process model [6]. And Hawkes process [7] and its variants can encapsulate the self-triggering and mutual triggering modes occurring among different complex event sequences in an unambiguous and quantitative fashion, so it is dominantly employed in a variety of application domain.\\
\indent Hawkes process can be divided into parametric [8,9] and nonparametric Hawkes process[10,11,12]. For neural Hawkes process, Du et al. proposed the Recurrent Marked Temporal Point Processes (RMTPP) [13,14]. Xiao et al. introduced the Intensity RNN [15, 16]. Mei et al. built the Neural Hawkes Process (NHP) [17].\\
\indent
In 2017, the transformer structure [18] has achieved good results, which has attracted extensive attention. This structure adopted by transformer completely gets rid of the other neural network structures, such as RNN, and CNN, and only puts attention mechanism to use for dealing with sequential tasks. In view of this, Zhang et al. devised Self Attention Hawkes Process (SAHP) [19], which takes advantage of the influence of historical events and finds the probability of the next event through self-attention mechanism. Zuo et al. presented the Transformer Hawkes Process (THP) [20], in which they integrated the transformer structure into the field of point process.\\
\indent
However, the previous Hawkes process model based on attention mechanism uses injudiciously the intrinsic event and time information existing in the asynchronous event sequence, but simply adds the encoding of event type conversion and the location encoding of time conversion to the encoder. Meanwhile, we speculate that the learner built from a single transformer may suffer from learning bias. motivated by ensemble learning [21] and Transformer-XL [22], we generate a tri-transformer Hawkes process model (Tri-THP), in which event and time information are added to dot product attention as auxiliary information recurrently to form a new multi-head attention. The results show that compared with the existing models, the performance of our Tri-THP model has been greatly improved. The main contributions of our study are as follows:\\
\indent
1) The proposed three different THPs: event type embedding THP (ETE-THP), the primary THP (PRI-THP), and temporal embedding THP (TE-THP) are fused into a seamless organic whole, which are related and complemented with each other positively, and are utilized to distil event type and temporal embedding information underlying in the asynchronous event sequences. \\
\indent
2) We carry out two novel dot-product attention operations, so far as we know, there were not any research reports about this.\\
\indent
3) Finally, The experimental results on both real-life and synthetic asynchronous event sequence have verified empirically the effect of our Tri-THP algorithm.
\section{Related Works}
In this section, we briefly recapitulate Hawkes Process [7], Transformer [18], Transformer Hawkes Process [20] and ensemble learning [21].\\
\indent
Hawkes process [7], which is belong to self-exciting process, its intrinsic mechanism allows that previous historical events have stimulating effects on the occurrence of future ones, and the cumulative influence of historical events is intensity superposition. The intensity function for Hawkes process is formulated as follows:
\begin{equation}
\lambda (t) = \mu (t) + \sum\limits_{i:{t_i} < t} {\psi (t - {t_i})} 
\end{equation}
where $\mu (t) $ is the base intensity. For the sake of simplicity and convenient calculation, or only involving the trigger mode, i.e., causalities among events, $\mu (t) $  is usually set to constant. $\psi ( \cdot )$  is a predefined decay function, also known as the influence function, in general, it is selected as the exponential decay function. From Eq.(1) ,we note  that every event in the Hawkes process has a stimulating effect on the occurring current event, and the stimulating effect reduces with time. However, this Hawkes process model does not involve inhibition effect between events, which is obviously inconsistent with the actual situation and impairs the expressiveness of the model.\\
\indent
In 2017, Google launched the transformer structure [18], which has achieved promise results and become a current mainstream model. This structure replaces the RNN, CNN and other network structures with attention mechanism to deal with sequential tasks. However, each fragment in the model is divided according to the length, and some relevant information may be lost when cutting, which will lead to context fragmentation. Then, motivated by this, Dai et al. proposed transformer XL [27]. This model uses the segment level recurrence mechanism to solve the fragmentation problem. The attention calculation in each layer needs to be encoded by relative position, which is about 450$\%$   longer than the original transformer learning dependency.\\
\indent
Inspired by the encoder structure in transformer [18], Zou et al. design the Transformer Hawkes process (THP) [20]. THP extends the transformer structure to a continuous time domain. It uses the encoder in transformer to obtain the hidden representation of sequence data, and uses the hidden representation to represent the continuous conditional intensity function, and obtains very good performance. However, in theory, attention can be associated with two words at any distance, but in practice, due to the limited computing resources, it is still unable to deal with very long input. Therefore, inspired by [27], we introduce event and time auxiliary information into attention calculation to further enhance learning dependence. \\
\indent
The goal of machine learning is to learn a stable model with good performance in all aspects, but in most cases, we can only get more than one model with good performance in some aspects. Ensemble learning is to combine multiple weak supervised models to get a better and more comprehensive strong supervised model. The potential idea of ensemble learning is to improve the prediction accuracy of the final result by combining multiple weaker models. Therefore, in order to improve the robustness and prediction accuracy of the model, we design three heterogeneous learners to learn the hidden representation respectively, and then weighted the hidden representation to get the final hidden representation.
\section{The Proposed Tri-THP Framework}
In this section, we will elaborate our Tri-THP model in detail. Suppose that we have an event sequence $X = \left\{ {\left( {{t_i},{k_i}} \right)} \right\}_{i = 1}^N$, in which each event has a type ${k_i} \in \left\{ {1,2, \ldots ,K} \right\}$. Each pair of event and time stamp $\left( {{t_i},{k_i}} \right)$  corresponds to an event of type ${k_i}$  which occurs at time ${t_i}$.
\subsection{Tri-Transformer Hawkes Process}
The core thought of our Tri-THP model is to build three THPs with different dot-product attention using different auxiliary information, and harvest diverse and heterogeneous side information from asynchronous event sequence. Before diving into the discussion of our Tri-THP model, we briefly present how to acquire the embedding of the occurrence time and type of the events, i.e., we need to transform the asynchronous event sequence to temporal and event encoding. Similar to [18] and [20], our temporal encoding expression is as follows:
\begin{equation}
[{\textbf{c}}(t_i )]_j  = \left\{ \begin{array}{l}
	\cos \left( {t_i /10000^{\frac{{j - 1}}{Z}} } \right),{\rm{if\;}}j{\rm{\;is\; odd,}} \\ 
	\sin \left( {t_i /10000^{\frac{j}{Z}} } \right),{\rm{if\;}}j{\rm{\;is\;even}}{\rm{.}} \\ 
\end{array} \right.
\end{equation}
where $\textbf{c}$  is the temporal encoding, $\textbf{c}\left( {{t_i}} \right) \in {\mathbb{R}^Z}$, $Z$ is the preassigned dimension,here $j \in \left[ {0, \ldots ,Z - 1} \right]$. That is to say, each component of temporal encoding denotes a sine curve. In terms of encoding event type, the unique one-hot encoding ${\textbf{k}_i} \in {\mathbb{R}^K}$  of each event type  is multiplied by the embedding matrix $\textbf{M} \in {\mathbb{R}^{Z \times K}}$  to get the event encoding $\textbf{M}{\textbf{k}_i} \in {\mathbb{R}^Z}$, so the sequence encoding of event sequence $X = \left\{ {\left( {{t_i},{k_i}} \right)} \right\}_{i = 1}^N$  is embedded as the sum of temporal encoding ${\textbf{C}^T}$  and event encoding ${\left( {\textbf{M}\textbf{Y}} \right)^T}$, where $\textbf{C} = \left[ {\textbf{c}\left( {{t_1}} \right), \ldots ,\textbf{c}\left( {{t_N}} \right)} \right] \in {\mathbb{R}^{Z \times N}}$, and $\textbf{Y} = \left[ {{\textbf{k}_1}, \ldots ,{\textbf{k}_N}} \right] \in {\mathbb{R}^{K \times N}}$. Therefore, each row in the final encoding corresponds to the temporal and event encoding in the sequence, respectively. Then, we feed the obtained sequence encoding into our Tri-THP model, and the schematic diagram of Tri-THP is shown in Fig. 1.\\
\indent
As shown in Fig. 1, we build three different learners: ETE-THP, PRI-THP, and TE-THP, which are complementary to each other, to carry out far more full-featured and diverse hidden state extraction. One that is shown on the left, we dub ETE-THP, is a learner that supplies with auxiliary information of event type embedding to multi-head attention module of the basic THP model; One that is depicted in the middle, we call PRI-THP, is the primary learner; One that is illustrated in the right, we call TE-THP, is a learner that equips with auxiliary information of temporal embedding to multi-head attention of the normal THP model. Three hidden states ${\bm{H}_1},{\bm{H}_2},{\bm{H}_3}$ are learned by three different learners respectively, and then the final hidden state is obtained by weighting the learned three hidden states with the three weights ${\lambda_1},{\lambda _2},{\lambda _3}$  obtained by training. In order to prevent overfitting and improve the robustness and performance of the model, we also incorporate dropout [23], layer normalization [24] and residual connection [25]. The specific training process of our Tri-THP model is shown in algorithm 1. \\
\indent
\begin{figure}[!htbp]
	\centering
	\includegraphics[width=\textwidth]{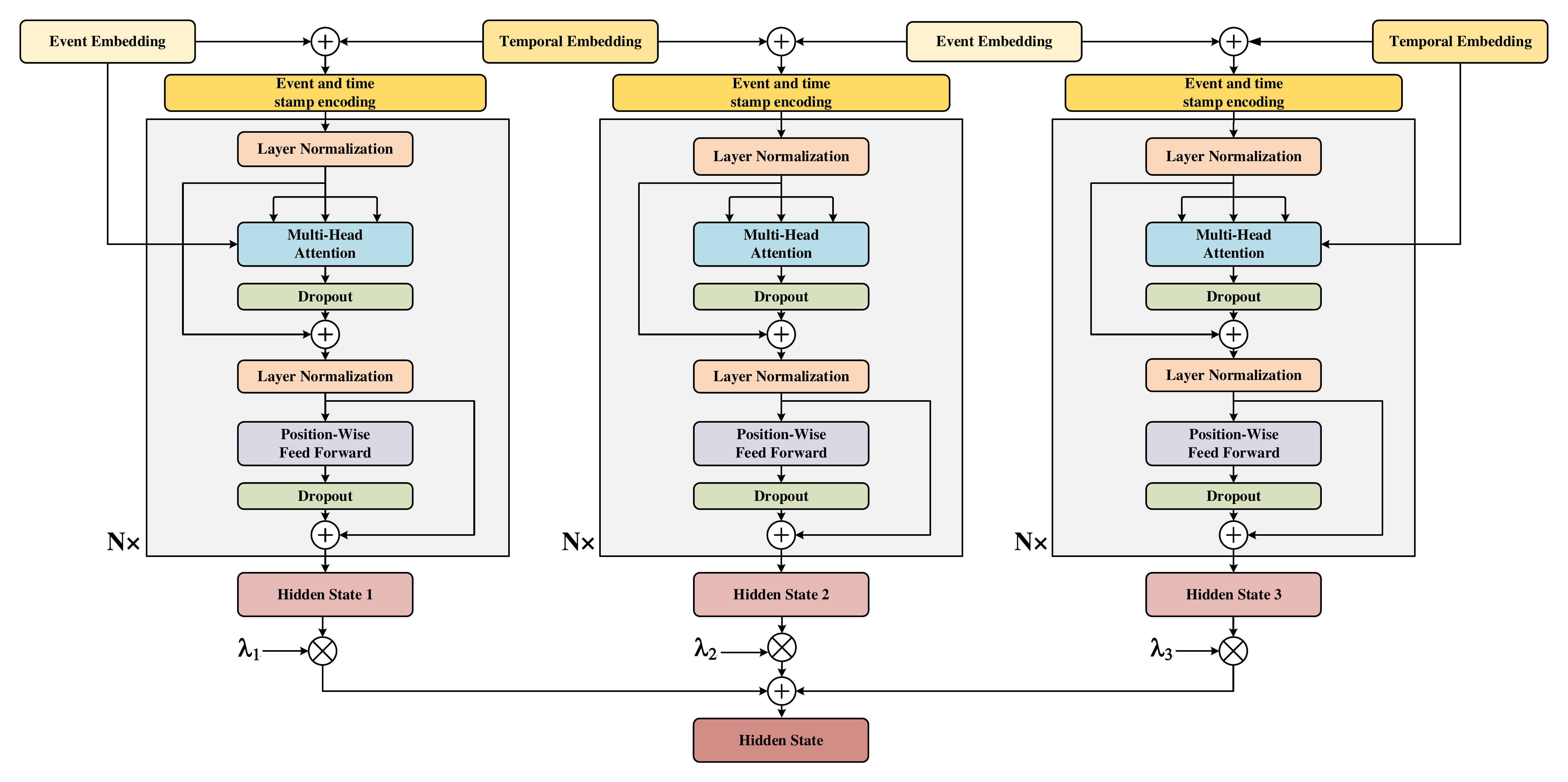}
	\caption{Diagrammatic sketch for our proposed Tri-THP model. Our Tri-THP comprises three modules: the THP module with auxiliary information of event type (left), the THP module without auxiliary information (middle), and the THP module with auxiliary information of time (right).} \label{fig1}
\end{figure}
\begin{algorithm}[!htbp]
	\caption{Tri-Transformer Hawkes Process (Tri-THP)}
	\label{alg1}
	\textbf{Input}: The number of encoding layers:$n$ , event type encoding ${\left( {\textbf{M}\textbf{Y}} \right)^T}$ , temporal encoding ${\textbf{C}^T}$ .\\
	\textbf{Output}: Hidden state of event sequence $\bm{H}\in\mathbb{R}^{Z\times N}$
	\begin{algorithmic}[1]
		\STATE Initialize state ${\bm{H}_1},{\bm{H}_2},{\bm{H}_3} \leftarrow {\left( {\textbf{M}\textbf{Y}} \right)^T}$		
		\FOR{$i$ in $n$}
		\STATE ${\bm{H}_1} \leftarrow {\bm{H}_1} + {\textbf{C}^T}$
		\STATE ${\bm{H}_2} \leftarrow {\bm{H}_2} + {\textbf{C}^T}$
		\STATE ${\bm{H}_3} \leftarrow {\bm{H}_3} + {\textbf{C}^T}$
		\STATE $\bm{H}_1 \leftarrow The\;i{\text{-th Encoding\_Layer}}(\bm{H}_1,{\left( {\textbf{M}\textbf{Y}} \right)^T}) $
		\STATE $\bm{H}_2 \leftarrow The\;i{\text{-th Encoding\_Layer}}(\bm{H}_2) $
		\STATE $\bm{H}_3 \leftarrow The\;i{\text{-th Encoding\_Layer}}(\bm{H}_3,{\textbf{C}^T}) $
		\ENDFOR
		\STATE $\bm{H} = {\lambda _1}{\bm{H}_1} + {\lambda _2}{\bm{H}_2} + {\lambda _3}{\bm{H}_3}$
		\STATE \textbf{return} $\bm{H}$
	\end{algorithmic}
\end{algorithm}
As shown in algorithm 1, in the initial stage, event encoding and temporal encoding are fed into the ETE-THP, PRI-THP, and TE-THP, respectively, and the input of each encoding layer is the output of the previous hidden layer plus temporal encoding. In our paradigm, we modify the conventional dot-product attention operation for multi-head attention in the encoding layer for different modules. In Fig. 1, the dot-product attention for the ETE-THP is written as follows:
\begin{equation}
\resizebox{.93\hsize}{!}{${\bm{A}}_1^s = Softmax\left[ {mask\left( {\frac{{\left( {{\bm{Q}}_1^s + {\bm{b}}_1^{sq}} \right){{\left( {{\bm{K}}_1^s} \right)}^T} + \left( {{\bm{Q}}_1^s + {\bm{b}}_1^{se}} \right){{\left( {{{\left( \textbf{MY} \right)}^T}{\bm{W}}_1^{even{t_s}}} \right)}^T}}}{{\sqrt {{Z_K}} }}} \right)} \right]{\bm{V}}_1^s$}	
\end{equation}
here we add the bias vector ${\bm{b}}_1^{sq}$  and  ${\bm{b}}_1^{se}$ to the query matrix ${\bm{Q}}_1^s$ ,respectively. Incorporating the bias vector ${\bm{b}}_1^{sq}$  and  ${\bm{b}}_1^{se}$ to the query matrix ${\bm{Q}}_1^s$ can strengthen model flexibility. The amended conventional dot-product attention operation is reflected in the item $\left( {{\bm{Q}}_1^s \!+\! {\bm{b}}_1^{sq}} \right){\left( {{\bm{K}}_1^s} \right)^T}$. The event information is introduced in the item  $\left( {{\bm{Q}}_1^s\! +\! {\bm{b}}_1^{se}} \right){\left( {{{\left( \textbf{MY} \right)}^T}{\bm{W}}_1^{even{t_s}}} \right)^T}$  to heighten the attention to the event type. Event encoding ${\left( {\textbf{M}\!\textbf{Y}\!} \right)^T}$ is linearly transformed into the item ${{{\left( \textbf{M\!Y} \right)}^T}\!{\bm{W}}_1^{even{t_s}}\!}$, and the  $s$-th attention head is linearly translated into matrix  ${\bm{W}}_1^{even{t_s}} \!\in\! {\mathbb{R}^{Z \times {Z_K}}}$. \\
\indent
The dot-product attention of the PRI-THP is expressed as follows:
\begin{equation}
{\bm{A}}_2^s = Softmax\left[ {mask\left( {\frac{{\left( {{\bm{Q}}_2^s + {\bm{b}}_2^{sq}} \right){{\left( {{\bm{K}}_2^s} \right)}^T}  }}{{\sqrt {{Z_K}} }}} \right)} \right]{\bm{V}}_2^s	
\end{equation}\\
\indent
On the basis of the THP model, we add a bias vector to the query matrix of dot-product attention, and derive the dot-product attention for the TE-THP:
\begin{equation}
\resizebox{.93\hsize}{!}{${\bm{A}}_3^s = Softmax\left[ {mask\left( {\frac{{\left( {{\bm{Q}}_3^s + {\bm{b}}_3^{sq}} \right){{\left( {{\bm{K}}_3^s} \right)}^T} + \left( {{\bm{Q}}_3^s + {\bm{b}}_3^{st}} \right){{\left( {{{\textbf{C}^T }}{\bm{W}}_3^{te{m_s}}} \right)}^T}}}{{\sqrt {{Z_K}} }}} \right)} \right]{\bm{V}}_3^s$}	
\end{equation}
i.e., the event information added in Eq. (3) is replaced with the time information in Eq. (5).\\
\indent
For each module, e.g., ETE-THP, PRI-THP, and TE-THP, the query, key and value matrix for the  $s$-th head attention are expressed as follows:
\begin{equation}
{{\bm{Q}}^s} = {\bm{HW}}_Q^s,{{\bm{K}}^s} = {\bm{HW}}_K^s,{{\bm{V}}^s} = {\bm{HW}}_V^s	
\end{equation}
here,  $\bm{H}$ is the input of each encoding layer as described in algorithm 1. ${\bm{W}}_Q^s \in {\mathbb{R}^{Z \times {Z_K}}}$ , ${\bm{W}}_K^s \in {\mathbb{R}^{Z \times {Z_K}}}$  and  ${\bm{W}}_V^s \in {\mathbb{R}^{Z \times {Z_V}}}$ are linear transformations of $\bm{H}$  respectively. The aim of mask operation is to ensure that future events in the matrix will not affect the attention weights of current events. In order to raise the expressiveness of the model, we utilize the multi-head attention of the following form:
\begin{equation}
{\bm{A}} = \left[ {{{\bm{A}}_1},{{\bm{A}}_2},...,{{\bm{A}}_S}} \right]{{\bm{W}}^{multi}}
\end{equation}
where ${{\bm{W}}^{multi}}$  is the aggregation matrix.\\
\indent
Then the attention weight matrix $\bm{A}$ is sent into position-wise feed-forward neural network to acquire the final hidden representation of the event sequence:
\begin{equation}
\bm{H} = {\text{ReLU}}({\bm{AW}}_1^{FC} + {{\bm{b}}_1}){\bm{W}}_2^{FC} + {{\bm{b}}_2}
\end{equation}\\
\indent
The above is construction procedure of the multi-head attention and feed-forward network for each module. After learning three modules, we will get three hidden matrices ${\bm{H}_1},{\bm{H}_2},{\bm{H}_3}$, and then multiply these three hidden matrices by the tradeoff coefficients ${\lambda_1},{\lambda _2}$ and ${\lambda _3}$  obtained from training, and then calculate weighted sum to obtain the final hidden state. 
\begin{equation}
\begin{gathered}
	\bm{H} = {\lambda _1}{\bm{H}_1} + {\lambda _2}{\bm{H}_2} + {\lambda _3}{\bm{H}_3} \hfill \\
	{\bm{h}}({t_i}) = \bm{H}\left( {i,:} \right) \hfill \\ 
\end{gathered} 
\end{equation}
where ${\bm{h}}({t_i})$ denotes the hidden state at time ${t_i}$.
\subsection{Conditional Intensity Function}
Conditional intensity function regulates the time point process. Similar to [20], we put the learned hidden state ${\bm{h}}({t_i})$ into the conditional intensity function:  
\begin{equation}
	\label{lam}
	\lambda_k ( {t\left| {\mathcal{H}_t } \right.} ) = f ({b_k } + { {\alpha _k \frac{{t - t_i }}
			{{t_i }}} + {\bm{w}_k^T \bm{h}( {t_i } )}})
\end{equation}
\indent
Then the whole strength function condition on ${\mathcal{H}_t }$ for $K$ types of sequences is formulated as follows:
\begin{equation}\label{all}
	\lambda( {t\left| {\mathcal{H}_t } \right.} ){\text{ = }}\sum\limits_{k = 1}^K {\lambda _k ( {t\left| {\mathcal{H}_t } \right.} )} 
	\vspace*{-4mm}
\end{equation}
\subsection{Loss Function for Forecasting Occurring Times and the Types of Events}
The predictive value for occurring times and the types of event can be deduced as follows:
\begin{equation}\label{nnpre}
	\begin{gathered}
		\hat {t}_{i + 1}  =  {\bm{W}^{time}\bm{h}( {t_i } )}\\
		\hat {\bm{p}}_{i + 1}  = Softmax( {\bm{W}^{type}\bm{h}( {t_i } )})\\
		\hat k_{i + 1}  = \mathop {\arg \max }\limits_k \hat {\bm{p}}_{i + 1} ( k )
	\end{gathered}
\end{equation}
where ${{\bm{W}}^{time}} \in {\mathbb{R}^{1 \times Z}}$  and  ${{\bm{W}}^{type}} \in {\mathbb{R}^{1 \times Z}}$ are the model parameters for occurring times and the types of event. More specifically, for sequence $X$, we utilize respectively the cross entropy loss of event type prediction and the square error of prediction occurring time of event to deduce the model parameters:
\begin{equation}
\begin{gathered}	
L_{time} (X) = \sum\nolimits_{i = 2}^{N} {( {t_i  - \hat t_i } )} ^2 \\
L_{type} (X) = \sum\nolimits_{i = 2}^{N} { - \textbf{k}} _i^T \log ( {\hat {\bm{p}_i} } )
\end{gathered}
\end{equation}
where the index $i$  begins at 2 to keep away from predicting the first event.
\subsection{Objective Function}
We are now at the position to elaborate the objective function. The logarithm likelihood for a given sequence $X$ can be inferred from Hawkes process theory:
\begin{equation}
	L( X) = {\sum\limits_{i = 1}^{N} {\log } \,\lambda ( {\left. {t_i } \right|\mathcal{H}_i } )} - {\int_{t_1 }^{t_{N} } {\lambda ( {\left. t \right|\mathcal{H}_t } )} dt}
\end{equation}\\
\indent
Provided that there are $L$  training sequences ${X_1},{X_2}, \ldots ,{X_L}$, according to the maximum likelihood estimation principle, we get
\begin{equation}
\max \sum\nolimits_{i = 1}^L {L({X_i})} 
\end{equation}\\
\indent
Unfortunately, directly using optimization approaches, such as the random gradient optimization algorithm Adam [26], to solve the problem (15) is unfeasible and unwise. The numeric expression for the second term $\rm{\Lambda} $$ = {\int_{t_1 }^{t_{N} } {\lambda ( {\left. t \right|\mathcal{H}_t } )} dt}$  in (14) cannot be obtained because the function $ {\lambda ( {\left. t \right|\mathcal{H}_t } )} $ is in the form of the deep neural network. Alternatively, the unbiased Monte Carlo integration method [27] and the biased numerical integration method [28] are incorporated to obtain the approximate value of $\rm{\Lambda}$. For the first approach the approximate expression for $\rm{\Lambda}$ is written as follows: 
\begin{equation}\label{monte}
	\hat \Lambda _{MC}  = \sum\limits_{i = 2}^N {\left( {t_i  - t_{i - 1} } \right)} ( {\frac{1}
		{O}\sum\limits_{o = 1}^O {\lambda ( {u_o } )} } )
\end{equation}
where ${u_o }$  is sampled from uniform distribution $U({t_{i - 1}},{t_i})$. The second one obtains the approximate value of $\rm{\Lambda}$ by using trapezoidal rule:
\begin{equation}\label{analy}
	\hat \Lambda _{NI}  = \sum\limits_{i = 2}^{N} {\frac{{t_i  - t_{i - 1} }}
		{2}} ( {\lambda ( {\left. {t_i } \right|\mathcal{H}_i } ) + \lambda ( {\left. {t_{i - 1} } \right|\mathcal{H}_{i- 1} } )} )
\end{equation}\\
\indent
Now, recall the Eq.(13), and sum the  three terms:  $L({X_i})$, ${L_{type}}({X_i})$,and   ${L_{time}}({X_i})$,we obtain the objective function  for our Tri-THP model : 
\begin{equation}
\min \sum\limits_{i = 1}^X { - L({X_i})}  + {L_{type}}({X_i}) + {L_{time}}({X_i})
\end{equation}
\section{Experimental Results}
In this section, we first introduce the details of the dataset, and then compare our model with the existing baseline approaches on synthetic and real-life datasets. We evaluate the model through log-likelihood (in nats), event prediction accuracies and root mean square errors for occurring time of events.
\subsection{Datasets}
In this subsection, we use two artificial datasets:Synthetic and Neural Hawkes[17], and four real-world datasets of event sequences:Electrical Medical Records[29], StackOverflow[30], Financial Transactions [14] and Retweets[31], to carry out experiments. Table 1 describes the characteristics of each dataset.
\begin{table}[!htbp]
	\centering 
	\caption{Characteristics of datasets used in experiments.}\label{tab1}
	\label{tb1}
	\begin{tabular}{c c c c c}
		\hline
		\multirow{2}{*}{Dataset} & \multirow{2}{*}{\textit{C}} & \multicolumn{3}{c}{Sequence Length} \\ \cline{3-5} 
		&                    & Min       & Aver.       & Max       \\ \hline
		Synthetic                & 5                  & 20        & 60          & 100       \\
		NeuralHawkes                & 5                  & 20        & 60          & 100       \\
		Retweets                 & 3                  & 50        & 109         & 264       \\
		MIMIC-II                 & 75                 & 2         & 4           & 33        \\
		StackOverflow            & 22                 & 41        & 72          & 736       \\
		Financial                 & 2                  & 829       & 2074        & 3319      \\ \hline
	\end{tabular}
\end{table}\\
\subsection{Experimental results and comparison}
We compare the performance of the Tri-THP model with that of the baseline model. First, we compare the log-likelihood value, which is the basic measure of the fitting degree of the model. Log-likelihood values of baseline and Tri-THP on different datasets are summed up in Table 2.\\
\indent
As shown in Table 2, the log-likelihood function values of Tri-THP on all test datasets are significantly better than the existing baselines. This shows that Tri-THP is more effective than the existing baselines in modeling event sequence. We think that this is because we use the idea of ensemble learning to design three learners. This way can effectively improve the stability and fitting similarity of the model, which is just the significance of the log-likelihood index. However, log-likelihood represents the similarity of the model, which has little effect on practical application.
\begin{table}[!htbp]
	\centering 
	\caption{The value of log-likelihood function on the test datasets for different models.}
	\label{333}
	\begin{tabular}{cccccc}
		\hline
		Datasets  & RMTPP            & NHP   & SAHP  & THP   & Tri-THP  \\ \hline
		Synthetic & \textbackslash{} & -1.33 &0.52 &0.834 & \textbf{6.036} \\
		NeuralHawkes & \textbackslash{} & -1.02 &0.241 &0.966 & \textbf{6.601} \\
		Retweets  & -5.99        & -5.06 & -5.85 & -4.69  & \textbf{2.611} \\
		StackOverflow  & -2.6    & -2.55 & -1.86 & -0.559  & \textbf{-0.544} \\
		MIMIC-II  & -1.35        & -1.38 & -0.52 & -0.143  & \textbf{-0.081} \\
		Financial  & -3.89       & -3.6 & \textbackslash{} & -1.388  & \textbf{-0.651} \\ \hline
	\end{tabular}
\end{table}\\
\indent
Therefore, we need to pay more attention to the prediction accuracies of event time and event type, which is of great practical significance. Through these two indicators, we can predict events in different fields and promote or prevent occurrences of unfavorable events. The prediction accuracies of event type are listed in Table 3.
\begin{table}[!htbp]
	\centering 
	\caption{Predict accuracies of different models on various datasets.}
	\label{333}
	\begin{tabular}{ccccc}
		\hline
		Datasets  & RMTPP            & NHP   & THP   & Tri-THP  \\ \hline
		StackOverflow & 45.9 & 46.3 & 46.79 & \textbf{46.81} \\
		MIMIC-II  & 81.2     & 83.2 & 83.2 & \textbf{84.1} \\
		Financial & 61.95    & 62.2 & 62.23  & \textbf{62.31}  \\ \hline
	\end{tabular}
\end{table}
\begin{figure}[!htbp]
	\centering
	\includegraphics[scale=0.17]{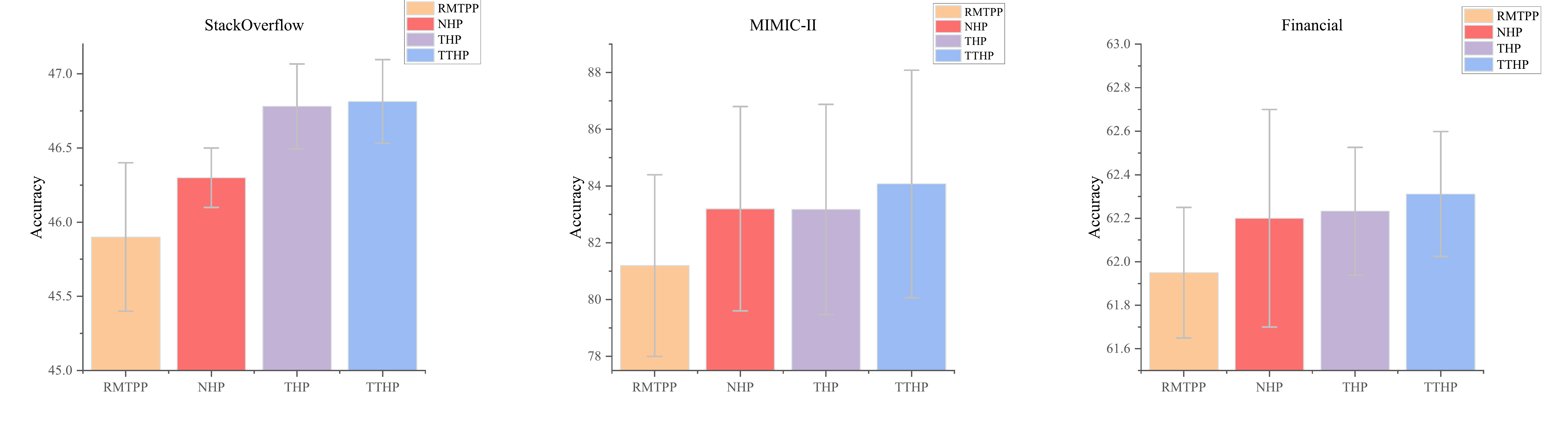}
	\caption{Prediction accuracies of baselines and Tri-THP. On the basis of five train-dev-test partitions, five experiments are carried out on each dataset, and the mean and standard deviation of different models are obtained.}
	\label{acc}
\end{figure}\\
\indent
It can be seen from table 3 and Fig. 2.  that the event prediction accuracies of our Tri-THP model is better than that of the existing baseline models on complex datasets. We speculate that it is caused by the learner with event auxiliary information. By adding event auxiliary information to multi-head attention, the learner can pay more attention to event types, Therefore, it can effectively improve the accuracies of event type prediction.\\
\indent
For the prediction of the occurrence time of various events, we use root mean square error (RMSE) as the unified measurement and evaluation creteria. The RMSE on different datasets for the baselines and our Tri-THP model are compared in Table 4.
\begin{table}[!htbp]
	\centering 
	\caption{RMSE of different models on various datasets.}
	\label{333}
	\begin{tabular}{ccccc}
		\hline
		Datasets  & RMTPP            & NHP   & THP   & Tri-THP  \\ \hline
		StackOverflow & 9.78 & 9.83 & 4.99 & \textbf{3.89} \\
		MIMIC-II  & 6.12     & 6.13 & 0.859 & \textbf{0.858} \\
		Financial & 1.56    & 1.56 & 0.02575  & \textbf{0.02550}  \\ \hline
	\end{tabular}
\end{table}\\
\indent
As shown in the Table 4 above, we can find that the RMSE of Tri-THP is higher than that of the baseline models. We guess that it is caused by the learner with time auxiliary information. By adding time auxiliary information to the multi-head attention, the learner can pay more attention to the time dependence relationships, so it can effectively reduce the root mean square error for occurrence time prediction. Our model shows some robustness on these datasets with obvious differences in sequence length and number of event types, which validates that the Tri-THP can effectively capture the short-term and long-term dependences between events.
\section{Conclusions and future work}
In this paper, we proposed a Tri-THP model. We introduce the event and time auxiliary information into the dot-product attention, and design three heterogeneous modules: ETE-THP, PRI-THP, and TE-THP. The experimental results verify that our Tri-THP model perform well on both real world and synthetic data, our Tri-THP model can effectively learn the complex short-term and long-term dependences.  Future works can consider not only improving the performance of the model, but also reducing the model complexity and the computational cost. Finally, we hope this work encourages further exploration of the interplay between and relative strong points of ensemble and attention mechanism.
%
%
%
%

\end{document}